\documentclass{article}

\usepackage{microtype}
\usepackage{graphicx}
\usepackage{subfigure}
\usepackage{booktabs} 

\usepackage{pifont}
\usepackage{multirow}
\usepackage{amsfonts}
\usepackage{amsmath}
\usepackage{longtable}
\usepackage{algorithm}
\usepackage{algorithmic}
\usepackage{bm}
\usepackage{makecell, boldline}
\usepackage{natbib}
\usepackage[dvipsnames]{xcolor}
\usepackage{color}
\usepackage{amssymb}
\usepackage{threeparttable}
\usepackage[justification=centering]{caption}

\definecolor{GrayGreen}{RGB}{125, 171, 171}

\DeclareMathOperator*{\argmax}{arg\,max}

\usepackage{hyperref}

\usepackage[accepted]{icml2021}

\icmltitlerunning{Self-Tuning for Data-Efficient Deep Learning}

\begin{document}

\twocolumn[
\icmltitle{Self-Tuning for Data-Efficient Deep Learning}



\icmlsetsymbol{equal}{*}

\begin{icmlauthorlist}
\icmlauthor{Ximei Wang}{equal,thu}
\icmlauthor{Jinghan Gao}{equal,thu}
\icmlauthor{Mingsheng Long}{thu}
\icmlauthor{Jianmin Wang}{thu}
\end{icmlauthorlist}


\icmlaffiliation{thu}{School of Software, BNRist, Tsinghua University, Beijing, China, 100084. E-mail: Ximei Wang (wxm17@mails.tsinghua.edu.cn)}

\icmlcorrespondingauthor{Mingsheng Long}{mingsheng@tsinghua.edu.cn}

\icmlkeywords{Deep Learning, Transfer Learning, Semi-Supervised Learning}

\vskip 0.3in
]



\printAffiliationsAndNotice{\icmlEqualContribution}  

\begin{abstract}
Deep learning has made revolutionary advances to diverse applications in the presence of large-scale labeled datasets. However, it is prohibitively time-costly and labor-expensive to collect sufficient labeled data in most realistic scenarios. To mitigate the requirement for labeled data, semi-supervised learning (SSL) focuses on simultaneously exploring both labeled and unlabeled data, while transfer learning (TL) popularizes a favorable practice of fine-tuning a pre-trained model to the target data. A dilemma is thus encountered: Without a decent pre-trained model to provide an implicit regularization, SSL through self-training from scratch will be easily misled by inaccurate pseudo-labels, especially in large-sized label space; Without exploring the intrinsic structure of unlabeled data, TL through fine-tuning from limited labeled data is at risk of under-transfer caused by model shift. To escape from this dilemma, we present \emph{Self-Tuning} to enable data-efficient deep learning by unifying the exploration of labeled and unlabeled data and the transfer of a pre-trained model, as well as a Pseudo Group Contrast (PGC) mechanism to mitigate the reliance on pseudo-labels and boost the tolerance to false labels. Self-Tuning outperforms its SSL and TL counterparts on five tasks by sharp margins, \emph{e.g.}~it doubles the accuracy of fine-tuning on \emph{Cars} with $15\%$ labels. 

\end{abstract}

\section{Introduction}
\label{intro}
In the last decade, deep learning has made revolutionary advances to diverse machine learning problems and applications in the presence of large-scale labeled datasets. However, in most real-world scenarios,  it is prohibitively time-costly and labor-expensive to collect sufficient labeled data through manual labeling, especially when labeling must be done by an expert such as a doctor in medical applications. To mitigate the requirement for labeled data, semi-supervised learning (SSL) focuses on simultaneously exploring both labeled and unlabeled data, while transfer learning (TL) popularizes a favorable practice of fine-tuning a pre-trained model to the target data. 

Semi-supervised learning (SSL) is a powerful approach for addressing the lack of labeled data by also exploring unlabeled examples. 
Recent advances in semi-supervised learning~\cite{sohn2020fixmatch, chen2020big} reveal that self-training~\cite{Lee13Pseudolabel}, which picks up the class with the highest predicted probability of a sample as its pseudo-label, is empirically and theoretically~\cite{wei2021theoretical} proved effective on unlabeled data. However, an obvious obstacle in pseudo-labeling is the \textit{confirmation bias}~\cite{arazo2020pseudolabeling}: the performance of a student is restricted by the teacher when learning from inaccurate pseudo-labels. In a prior study, we investigated the current state-of-the-art SSL method, FixMatch~\cite{sohn2020fixmatch}, on a target dataset \textit{CUB-200-2011}~\cite{WahCUB_200_2011} containing $200$ bird species. As Figure~\ref{fig:label_size} shows, keeping the same label proportion of 15\%, the test accuracy of FixMatch drops rapidly as the descending accuracy of pseudo-labels when the label space enlarges from $10$ (\textit{CUB10}) to $200$ (\textit{CUB200}). This observation reveals that
SSL through self-training from scratch, without a decent pre-trained model to provide an implicit regularization, will be easily misled by inaccurate pseudo-labels, especially in large-sized label space.

\begin{figure*}[!htbp]
	\centering
	\subfigure[\textbf{Transfer Learning}]{
		\includegraphics[width=0.23\textwidth]{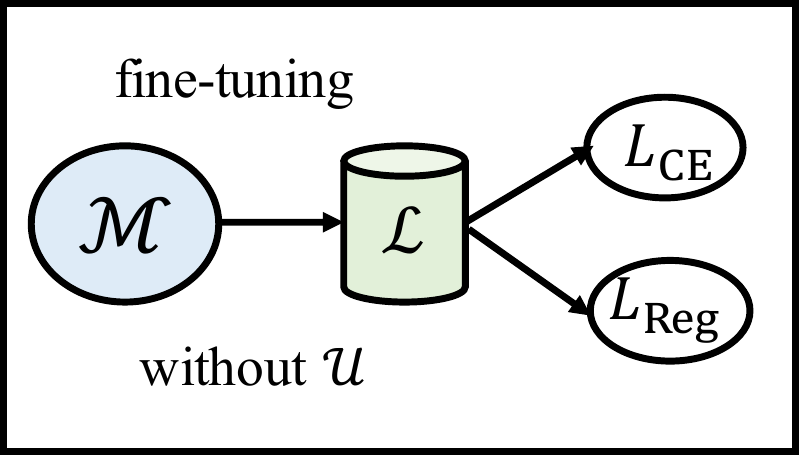}
		\label{fig:TL}
	}
	\subfigure[\textbf{Semi-supervised Learning}]{
		\includegraphics[width=0.23\textwidth]{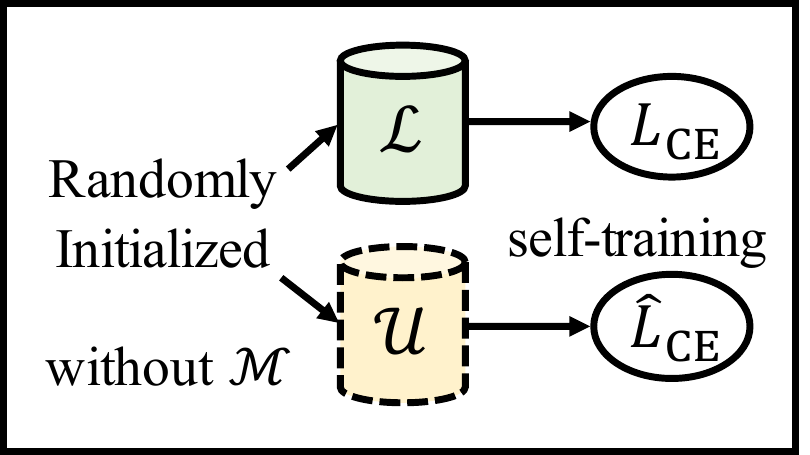}
		\label{fig:SSL}
	}
	\subfigure[\textbf{SimCLRv2}]{
		\includegraphics[width=0.23\textwidth]{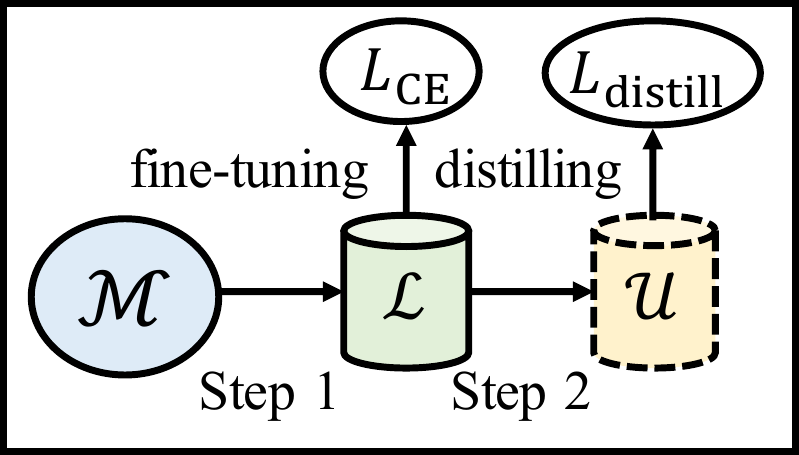}
		\label{fig:fine_tune_and_distill}
	}
	\subfigure[\textbf{Self-Tuning} (ours)]{
		\includegraphics[width=0.23\textwidth]{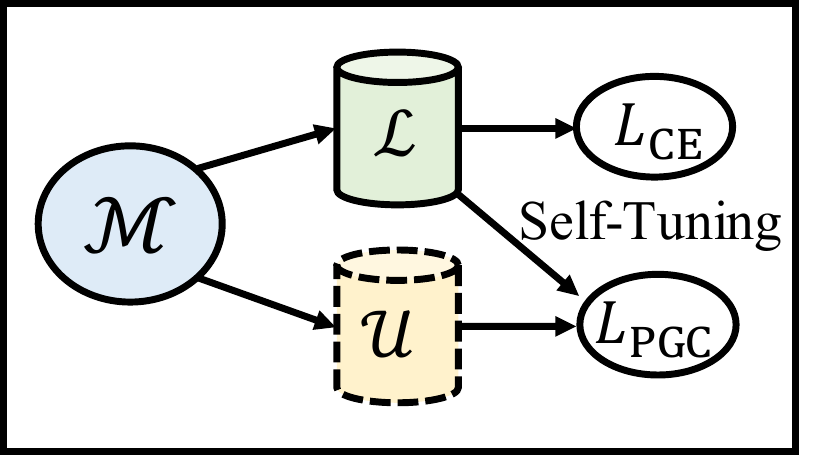}
		\label{fig:Self-Tuning}
	}
	\caption{Comparisons among techniques. (a) \textbf{Transfer Learning}: only fine-tuning on $\mathcal{L}$ with a regularization term; (b) \textbf{Semi-supervised Learning}: a common practice for SSL is a CE loss on $\mathcal{L}$ while self-training on $\mathcal{U}$ without a decent pretrained model; (c) \textbf{SimCLRv2}: fine-tune model $\mathcal{M}$ on $\mathcal{L}$ first and then distill on $\mathcal{U}$; (d) \textbf{Self-Tuning}: unify the exploration of $\mathcal{L}$ and $\mathcal{U}$ and the transfer of model $\mathcal{M}$.}
	\label{fig:paradigms}
	\vspace{-5pt}
\end{figure*}%

Fine-tuning a pre-trained model to a labeled target dataset is a popular form of transfer learning (TL) and increasingly becoming a common practice within computer vision (CV) and natural language processing (NLP) communities.
For instance, ResNet~\cite{he2016deep} and EfficientNet~\cite{Tan19EfficientNet} models pre-trained on ImageNet~\cite{deng2009imagenet} are widely fine-tuned into various CV tasks, while BERT~\citep{devlin2018bert} and GPT-3~\cite{brown2020language} models pre-trained on large-scale corpus achieve strong performance on diverse NLP tasks. Recent works on fine-tuning mainly focus on how to better exploit a target labeled data and a pre-trained model from various perspectives, such as weights~\cite{Li2018L2SP}, features~\cite{Li2019Delta}, singular values~\cite{Chen2019BSS} and category relationship~\cite{you2020co}. In a prior study, we investigated the current state-of-the-art TL method,  Co-Tuning, on standard TL benchmarks: \textit{CUB-200-2011}  and \textit{Stanford Cars}~\cite{KrauseStarkDengFei-Fei_3DRR2013}.
As shown in Figure~\ref{fig:label_proportion}, the test accuracy of Co-Tuning declines rapidly as the number of labeled data decreases. This observation tells us: without exploring the intrinsic structure of unlabeled data, TL through fine-tuning from limited labeled data is at risk of under-transfer caused by \textit{model shift}: the fine-tuned model shifts towards the limited labeled data and leaves away from the original smooth model pre-trained on a large-scale dataset, causing an unsatisfactory performance on the test set.

Realizing the drawback of only developing TL or SSL technique, a recent state-of-the-art paper named SimCLRv2~\cite{chen2020big} provided a new and interesting solution by fine-tuning from a big ImageNet pre-trained model $\mathcal{M}$ on a labeled data $\mathcal{L}$ first and then distilling on the unlabeled data $\mathcal{U}$. 
Its effectiveness has been demonstrated when fine-tuning to the same ImageNet dataset. However, we empirically found its unsatisfactory performance when transferring to \textit{cross-domain} datasets, especially in the low-data regime, as reported in Table~\ref{table_TL_benchmark}.  We hypothesize that the \textit{sequential} form between first fine-tuning on $\mathcal{L}$ and then distilling on $\mathcal{U}$ that SimCLRv2 adopts
is to blame, since the fine-tuned model would easily shift towards the limited labeled data with sampling bias and leaves away from the original smooth model pre-trained on a large-scale dataset.

To escape from the dilemma, we present \emph{Self-Tuning}, a novel approach to enable data-efficient deep learning.
Specifically, to address the challenge of \textit{confirmation bias} in self-training, a Pseudo Group Contrast (PGC) mechanism is devised to mitigate the reliance on pseudo-labels and boost the tolerance to false labels, after realizing the drawbacks of cross-entropy (CE) loss and contrastive learning (CL) loss. The model trained by CE loss will be easily confused by false pseudo-labels since it focuses on learning a hyperplane for discriminating each class from the other classes, while standard CL loss lacks a mechanism to tailor pseudo-labels into model training, leaving the useful discriminative information on the shelf. 
Further, we propose to unify the exploration of labeled and unlabeled data and the transfer of a pre-trained model to tackle the \textit{model shift} problem, different from the sequential form of exploring labeled and unlabeled data. Comparisons among these techniques are shown in Figure~\ref{fig:paradigms}, revealing the advantages of Self-Tuning.

In summary, this paper has the following contributions:
\begin{itemize}
	\item Realizing the dilemma of TL and SSL methods that only focus on either the pre-trained model or unlabeled data, we unleash the power of both worlds by proposing {a new setup} named data-efficient deep learning.
	\item To tackle model shift and confirmation bias problems, we propose \emph{Self-Tuning} to unify the exploration of labeled and unlabeled data and the transfer of a pre-trained model, as well as a general Pseudo Group Contrast mechanism to mitigate the reliance on pseudo-labels and boost the tolerance to false labels.
	\item Comprehensive experiments demonstrate that \textit{{Self-Tuning}} outperforms its SSL and TL counterparts on five tasks by sharp margins, \emph{e.g.}~it doubles the accuracy of fine-tuning on \emph{Cars} with $15\%$ labels. 
\end{itemize}

\section{Related Work}

\subsection{Self-training in Semi-supervised Learning}
Self-training~\cite{Yarowsky95Unsupervised, Bengio05entropy, Lee13Pseudolabel} is a widely-used technique for exploring unlabeled data with deep neural networks, especially in SSL. Among techniques of self-training, pseudo-labeling~\cite{Lee13Pseudolabel} is one of the most popular forms by leveraging the model itself to obtain artificial labels for unlabeled data. Recent advances in SSL reveal that self-training is empirically~\cite{sohn2020fixmatch} and theoretically~\cite{wei2021theoretical} effective on unlabeled data.
These methods either require stability of predictions under different data augmentations~\cite{Tarvainen17MeanTeacher, xie2019unsupervised, sohn2020fixmatch} (also known as input consistency regularization) or fit the unlabeled data on its predictions generated by a previously learned model~\cite{Lee13Pseudolabel, chen2020big}. Specifically, UDA~\cite{xie2019unsupervised} reveals that the quality of noising produced by advanced data augmentation methods plays a crucial role in SSL. FixMatch~\cite{sohn2020fixmatch} uses the model’s predictions on weakly-augmented unlabeled images to generate pseudo-labels for the strongly-augmented versions of the same images. A recent state-of-the-art paper named SimCLRv2~\cite{chen2020big} provided a new solution for SSL by first fine-tuning from the labeled data and then distilling on the unlabeled data.

However, without a decent pre-trained model to provide an implicit regularization, SSL through self-training from scratch will be {easily misled by inaccurate pseudo-labels}, especially in large-sized label space.
Meanwhile, an obvious obstacle in pseudo-labeling is \textit{confirmation bias}~\cite{arazo2020pseudolabeling}: the performance of a student is restricted by the teacher when learning from inaccurate pseudo-labels.

\subsection{Fine-tuning in Transfer Learning}
Fine-tuning a pre-trained model to a labeled target dataset is a popular form of transfer learning (TL) and widely applied in various applications. Previously, \citet{Donahue14DeCAF, OquabBLS14TransferMid} show that transferring features extracted by pre-trained AlexNet model to downstream tasks provides better performance than that of hand-crafted features. Later, \citet{yosinski2014transferable, AgrawalGM14Analyzing, GirshickDDM14} reveal that fine-tuning pre-trained networks work better than fixed pre-trained representations. Recent works on fine-tuning mainly focus on how to better exploit the discriminative knowledge of labeled data and the information of pre-trained models from different perspectives. (a) \textbf{weights}: L2-SP~\cite{Li2018L2SP} explicitly promotes the similarity of the final solution with pre-trained weights by a simple L2 penalty. (b) \textbf{features}: DELTA~\cite{Li2019Delta} constrains a subset of feature maps with the pre-trained activations that are precisely selected by channel-wise attention. (c) \textbf{singular values}: BSS~\cite{Chen2019BSS} penalizes smaller singular values to suppress untransferable spectral components to avoid negative transfer. (d) \textbf{category relationship}: Co-Tuning~\cite{you2020co} learns the relationship between source categories and target categories from the pre-trained model to enable a full transfer. Even when the target dataset is very dissimilar to the pre-trained dataset and fine-tuning brings no performance gain~\cite{RaghuZKB19Transfusion}, it can accelerate the convergence speed~\cite{HeGD19}.
Meanwhile, NLP research on fine-tuning has an alternative focus on resource consumption~\cite{HoulsbyGJMLGAG19NLP, Garg20SimppleTran}, selective layer freezing~\cite{ Wang19Totune}, different learning rates~\cite{SunQXH19FinetunBERT} and scaling up language models~\cite{brown2020language}.

However, without exploring the intrinsic structure of unlabeled data, TL through fine-tuning from limited labeled data is at risk of under-transfer caused by \textit{model shift}: the fine-tuned model shifts towards the limited labeled data after leaving away from the original smooth model pre-trained on a large-scale dataset, causing an unsatisfactory test accuracy on the target dataset that we concern.

\begin{figure*}[!htbp]
	\centering
	\includegraphics[width=1.0\linewidth]{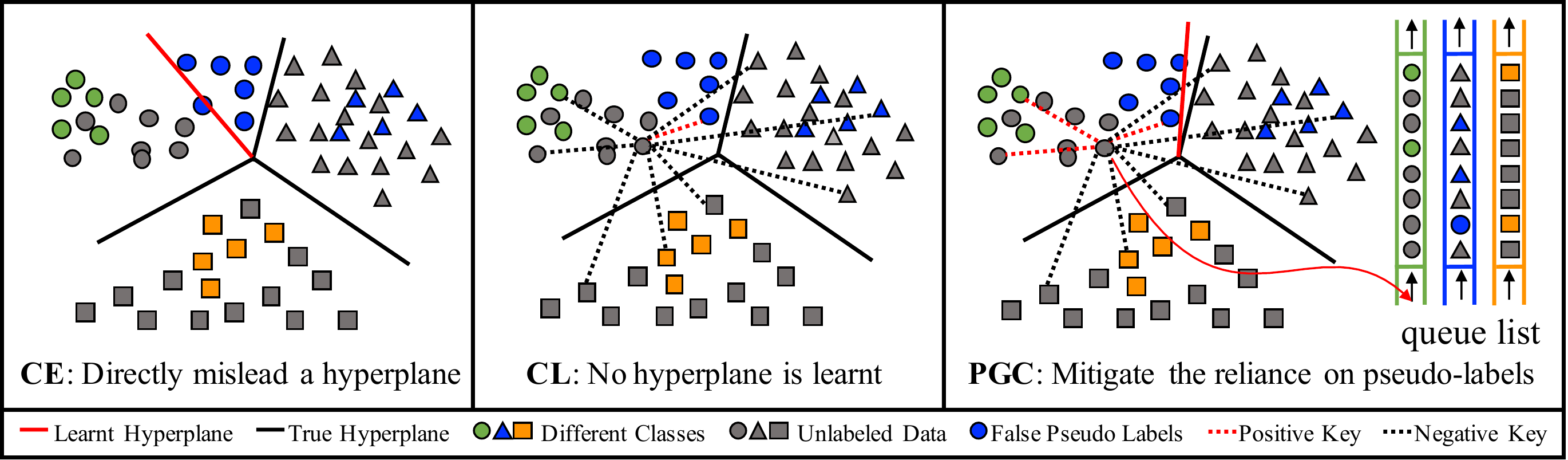}
	\caption{Comparison of various loss functions: (a) \textbf{CE}: cross-entropy loss will be easily misled by false pseudo-labels; (b) \textbf{CL}: contrastive learning loss underutilizes labels and pseudo-labels; (c) \textbf{PGC}:  Pseudo Group Contrast mechanism to mitigate confirmation bias.}
	\label{fig:losscomparison}
\end{figure*}

\section{Preliminaries}
\label{Preliminaries}
\subsection{The Devil Lies in Cross-Entropy Loss}
To figure out the \emph{confirmation bias} of pseudo-labeling, we first delve into the standard cross-entropy (CE) loss that most self-training methods adopt. Given labeled data
$\mathcal{L}$ with $C$ categories,  $y_i$ is the ground-truth label for each data point $\mathbf{x}_i$ whose prediction probability $\mathbf{p}_i = \mathcal{M}(\mathbf{x}_i )$ is generated from model $\mathcal{M}$. For each data point $\mathbf{x}_i$, the standard CE loss can be formalized as
\begin{equation}
\label{CE_loss}
L_{\rm CE} = - \sum_{c=1}^C \mathbf{1}({y} _ { i } =c)  \log \mathbf{p}_i^c ,
\end{equation}
where $\mathbf{1}(\cdot)\in\{0, 1\}$ is an indicator function that values $1$ if and only if the input condition holds. Similarly, for each data point $\mathbf{x}_i$ with prediction probability $\mathbf{p}_i = \mathcal{M}(\mathbf{x}_i )$, self-training loss on unlabeled data $\mathcal{U}$ is 
\begin{equation}
\label{pseudo_CE_loss}
\widehat{L}_{\rm CE} = - \sum_{c=1}^C \mathbf{1}(\widehat{y} _ { i } =c)  \mathbf{1}(z_i > t) \log \mathbf{p}_i^c,
\end{equation}
where $\widehat{y} _ { i } = \argmax_c \mathbf{p}_i^c$ is the pseudo-label for the input $\mathbf { x } _ { i }$ generated by a previously-learned model or from the input with different data augmentation, $z_i  = \max_c(\mathbf{p}_i^c) $ is the corresponding confidence, and $t$ is the threshold to select out more confident pseudo-labels. Note that the {confidence-threshold} $t$ is necessary in most self-training methods and set with a high value, \textit{e.g.} $t =0.95$ in FixMatch, or even with a complicated curriculum strategy. Such a self-training loss is effective in exploring unlabeled data. However,  as shown in Figure~\ref{fig:losscomparison}, the model trained by CE loss will be easily confused by false pseudo-labels since it focuses on learning a hyperplane for discriminating each class from the other classes, causing the unsatisfactory performance on target dataset with large-sized label space.

\subsection{Contrastive Learning Loss Underutilizes Labels}
To overcome  the drawbacks of \textit{class discrimination} for self-training, recent advanced researches of \textit{instance discrimination}~\citep{oord2018representation,wu2018unsupervised, he2019momentum, chen_simple_2020} attract our great attention. Given an encoded query $\mathbf{q}$ and encoded keys $\{{ \mathbf{k}_0},\mathbf{k}_1,\mathbf{k}_2,\cdots,\mathbf{k}_{D}\}$ with size $(D+1)$, a general form of contrastive learning (CL) loss with similarity measured by dot product for each data point on unlabeled data $\mathcal{U}$ is
\begin{equation}
\label{CL_loss}
L_{ \rm{CL}}=  -\log\frac{\exp(\mathbf{q}\cdot \mathbf{k}_{0}/\tau)}{\exp(\mathbf{q}\cdot \mathbf{k}_{0}/\tau) +  \sum_{d=1}^{D}\exp(\mathbf{q}\cdot \mathbf{k}_{d}/\tau)},
\end{equation}
where $\tau$ is a hyper-parameter for temperature scaling. Note that $\mathbf{k}_{0}$ is the only positive key that $\mathbf{q}$ matches since they are extracted from differently augmented views of the \textit{same} data example, while negative keys $\{\mathbf{k}_1,\mathbf{k}_2,\cdots,\mathbf{k}_{D}\}$ are selected from a dynamic queue which iteratively and progressively replace the oldest samples by the newly-generated keys. A contrastive loss maximizes the similarity between the query $\mathbf{q}$ with its corresponding positive key $\mathbf{k}_{0}$. According to the properties of the \textit{softmax} function adopted in Eq.~\eqref{CL_loss}, the similarity between the query with those negative keys $\{\mathbf{k}_1,\mathbf{k}_2,\cdots,\mathbf{k}_{D}\}$ is minimized.
Maximizing agreement between differently augmented views of the same data point,  CL loss focuses on exploring the intrinsic structure of data and is naturally independent of false pseudo-labels.
However, standard CL loss lacks a mechanism to tailor labels and pseudo-labels into model training, leaving the useful discriminative information on the shelf.

\section{Self-Tuning}
In data-efficient deep learning,  a pre-trained model $\mathcal{M}$, a labeled dataset $\mathcal{L} = \left\{ \left( \mathbf { x } _ i^L , {y} _i^L \right) \right\} _ { i = 1 } ^ { n _ { L} }$ and an unlabeled dataset $\mathcal{U} = \left\{ \left( \mathbf { x } _ { i }^U \right) \right\} _ { i = 1 } ^ { n _ { U } }$ in the target domain are given. Instantiated as a deep network, $\mathcal{M}$ is composed of a pre-trained backbone $f_0$ for feature extraction and a pre-trained head $g_0$, while fine-tuned ones are denoted by $f$  and $g$ respectively. $f$ is usually initialized as $f_0$ while $g$ is randomly initialized, since the target dataset usually has a different label space with size $C$ from that of pre-trained models. There are two obstables in such a practical paradigm: \textit{confirmation bias} and \textit{model shift}, which are addressed by pseudo group contrast mechanism and unifying the exploration respectively.
 
\subsection{Confirmation Bias: Pseudo Group Contrast }
\label{PGC_section}
As mentioned in Preliminaries~\ref{Preliminaries}, neither cross-entropy loss nor contrastive learning loss is a suitable loss function to address the challenge of confirmation bias in self-training. In this paper, a novel Pseudo Group Contrast (PGC) mechanism is raised to mitigate the reliance on pseudo-labels and boost the tolerance to false labels. Different from the standard CL which involves just a positive key in each contrast, PGC introduces \textit{a group of positive keys in the same pseudo-class} to contrast with all negative keys from other pseudo-classes. Specifically, for each data point $\mathbf{x}_i^U$ in unlabeled dataset $\mathcal{U}$, an encoded query $\mathbf{q}_i^U = h(f({\rm aug}_1(\mathbf{x}_i^U)))$ and an encoded key $\mathbf{k}_i^U = h(f({\rm aug}_2(\mathbf{x}_i^U)))$ are generated by a feature extractor $f$ following with a projector head $h$ on two differently-augmented views ${\rm aug}_1$ and ${\rm aug}_2$ of the same data example. By forwarding into the classifier $g$, a pseudo-label $\widehat{y}_i^U = \argmax_c g(f({\rm aug}_1(\mathbf{x}_i^U)))$ is attained.

For clarity, we focus on a particular data example  $\mathbf{ x }$ with pseudo-label $\widehat{y}$ and omit the subscript $i$ and the superscript $U$. Different from standard CL loss, a group of positive keys $\{\mathbf{k}_1^{\widehat{y} },\mathbf{k}_2^{\widehat{y} },\cdots,\mathbf{k}_{D}^{\widehat{y} }\}$ are selected according to its pseudo-label $\widehat{y}$, as well as its ${ \mathbf{k}_0^{\widehat{y} }}$ generated by its differently-augmented view. In this way, the scope of positive keys is successfully expanded from a {single one} to \textit{a group of instances} with size $D+1$. Complementarily, all keys from other pseudo-classes are seen as negative keys with size $[D\times (C-1)]$, selected from the dynamic queue list with size $[D\times C]$ according to their pseudo-labels. Note that, $D$ in PGC is equal to \textit{the queue size in standard CL divided by $C$}, resulting in a comparable memory consumption. Formally, for each data point $\mathbf{x}_i^U$ on unlabeled data $\mathcal{U}$, PGC loss is summarized as
\begin{equation}
\begin{aligned}
\label{PGC_loss}
\widehat{L}_{\text{PGC}} & =  - \frac{1}{D+1}\sum_{d=0}^{D} \log\frac{\exp(\mathbf{q}\cdot \mathbf{k}_{d}^{\widehat{y} }/\tau)}{\rm Pos + \rm Neg} \\
{\rm Pos} & = \exp(\mathbf{q}\cdot \mathbf{k}_{0}^{\widehat{y} }/\tau) +\sum_{j=1}^{D}\exp(\mathbf{q}\cdot \mathbf{k}_{j}^{\widehat{y}}/\tau) \\
{\rm Neg} & = \sum_{c=1}^{\{1,2,\cdots,C\} \backslash \widehat{y}} \sum_{j=1}^{D}\exp(\mathbf{q}\cdot \mathbf{k}_{j}^c/\tau),
\end{aligned}
\end{equation}

where the term of $\rm Pos$ denotes positve keys from the same pseudo-class $\widehat{y}$ while the term of $\rm Neg$ denotes negative keys from other pseudo-classes $\{1,2,\cdots,C\} \backslash \widehat{y}$. Obviously, PGC maximizes the similarity between the query $\mathbf{q}$ with its corresponding group of positive keys $\{{ \mathbf{k}_0^{\widehat{y} }}, \mathbf{k}_1^{\widehat{y} },\mathbf{k}_2^{\widehat{y} },\cdots,\mathbf{k}_{D}^{\widehat{y} }\}$ from the same pseudo-class $\widehat{y}$.

Further, according to the property of the \textit{softmax} function which generates a predicted probability vector with a sum of $1$, positive keys $\{{ \mathbf{k}_0^{\widehat{y} }}, \mathbf{k}_1^{\widehat{y} },\mathbf{k}_2^{\widehat{y} },\cdots,\mathbf{k}_{D}^{\widehat{y} }\}$ from the same pseudo-class will compete with each other. Therefore, {if some pseudo-labels in the positive group are wrong, those keys with true pseudo-labels will win this instance competition, since their representations are more similar to the query, compared to that of false ones}. Consequently, the model trained by PGC will be mainly updated by gradients of true pseudo-labels and largely avoid being misled by false pseudo-labels. Since PGC itself can mitigate the reliance on pseudo-labels and boost the tolerance to false labels, \textit{no confidence-threshold hyper-parameter $t$ is included in PGC}, making it easier to apply into new datasets than standard self-training in Eq.\eqref{pseudo_CE_loss}.
A conceptual comparison between PGC with CE and CL is shown in Figure~\ref{fig:losscomparison}. Ablation studies in Table~\ref{ablation_study} also confirm that PGC performs much better than CE and CL when initializing from an identical pre-trained model with the same pseudo-label accuracy.

\begin{figure}
	\centering
	\includegraphics[width=1.0\linewidth]{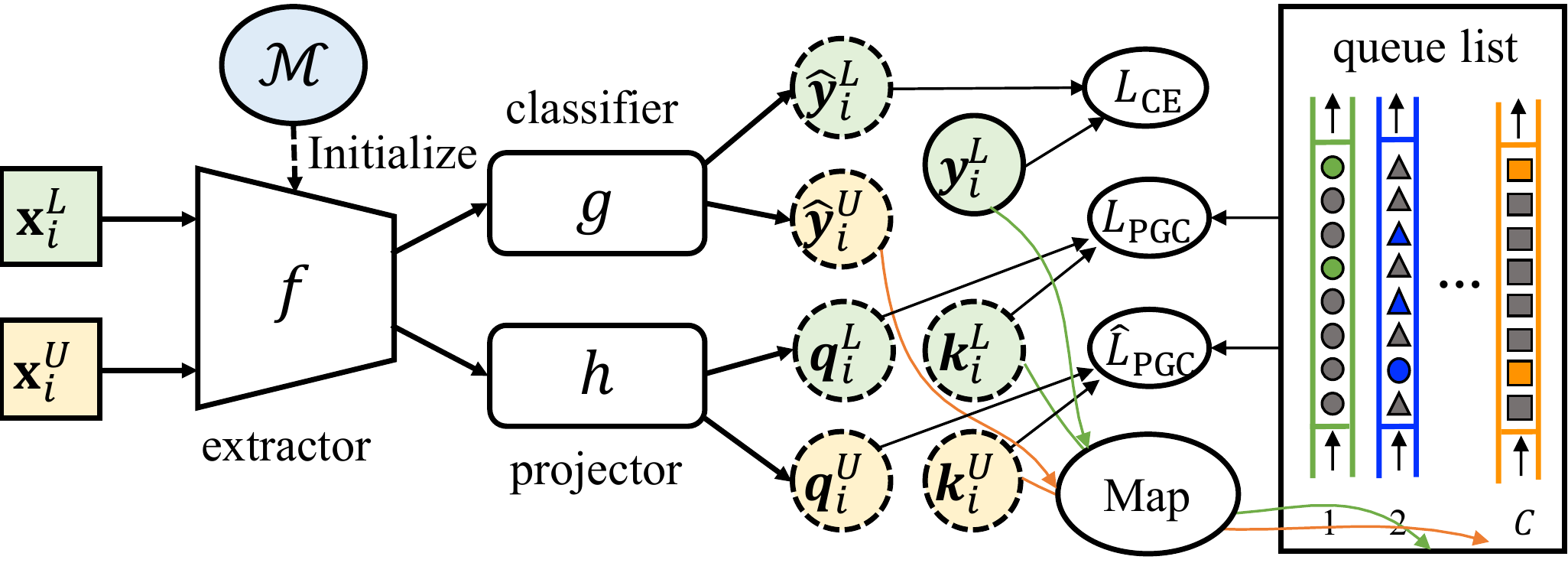}
	\caption{The network architecture of Self-Tuning. The ``\textit{Map}'' denotes a mapping function which assigns a newly-generated key to the corresoping queue according to its label or pseudo-label.}
	\label{fig:arch}
\end{figure}

\subsection{Model Shift: Unifying and Sharing}

Recall the \textit{model shift} problem of transfer learning through fine-tuning from limited labeled data: the fine-tuned model shifts towards the limited labeled data after leaving away from the original smooth model pre-trained on a large-scale dataset, causing an unsatisfactory performance on the test set. A recent state-of-the-art paper named SimCLRv2~\cite{chen2020big} gives an interesting solution of fine-tuning from a big pre-trained model $\mathcal{M}$ on a labeled data $\mathcal{L}$ first and then distilling on the unlabeled data $\mathcal{U}$. However, due to the sequential form it adopts, the fine-tuned model still easily shifts towards the limited labeled data with sampling bias and leaves away from the original smooth model. To this end, we propose to unify the exploration of labeled and unlabeled data and the transfer of a pre-trained model.

\paragraph{A unified form to fully exploit $\mathcal{M}$, $\mathcal{L}$ and $\mathcal{U}$}
Realizing the drawbacks of the sequential form of first fine-tuning on the labeled data and then distilling on the unlabeled data, we propose a unified form to fully exploit $\mathcal{M}$, $\mathcal{L}$ and $\mathcal{U}$ to tackle the model shift problem. First, initialized from a decently accurate pre-trained model, Self-Tuning has \textit{a better starting point to provide an implicit regularization} than the model trained from scratch on the target dataset. 

Further, the knowledge of the pre-trained model \textit{parallelly} flows into both the labeled and unlabeled data, which is different from the sequential form that overfits the limited labeled data first. Meanwhile, the parameters of the model will be simultaneously updated by gradients from both the labeled data $\mathcal{L}$ and unlabeled data $\mathcal{U}$. By exploring the label information of $\mathcal{L}$ and intrinsic structure of $\mathcal{U}$ at the same time in a unified form as shown in Figure~\ref{fig:Self-Tuning}, the model shift challenge is expected to be alleviated.

\paragraph{A shared queue list across $\mathcal{L}$ and $\mathcal{U}$}
Given a labeled data $\mathcal{L} = \left\{ \left( \mathbf { x } _ i^L , {y} _i^L \right) \right\} _ { i = 1 } ^ { n _ { L} }$ from $C$ categories, its ground-truth labels are readily-available.  For a data sample $\left( \mathbf { x }_i^L, {y}_i^L\right)$ in $\mathcal{L}$, its encoded query $\mathbf{q}_i^L = h(f({\rm aug}_1(\mathbf{x}_i^L)))$ and encoded key $\mathbf{k}_i^L = h(f({\rm aug}_2(\mathbf{x}_i^L)))$ are generated similarly. 
For clarity, we focus on a particular data example $\left( \mathbf { x }, {y}\right)$ and omit the subscript $i$ and the superscript $L$. Intuitively, we can simply replace the $\widehat{y}$ in Eq.~\eqref{PGC_loss} with ${y}$ to attain the ground-truth version of PGC on the labeled data. Formally, for each data point $\left( \mathbf { x }, {y}\right)$ on $\mathcal{L}$, PGC loss is summarized as
\begin{equation}
\begin{aligned}
\label{GGC_loss}
L_{\text{PGC}} & =  - \frac{1}{D+1}\sum_{d=0}^{D} \log\frac{\exp(\mathbf{q}\cdot \mathbf{k}_{d}^{{y} }/\tau)}{\rm Pos + \rm Neg},
\end{aligned}
\end{equation}
where the term of $\rm Pos$  and $\rm Neg$ are simlarly defined as that in Eq.~\eqref{PGC_loss} except replacing $\widehat{y}$ with ${y}$.

It is noteworthy that the queue list is \textit{shared} across labeled and unlabeled data, that is, encoded keys generated from both $\mathcal{L}$ and $\mathcal{U}$ will iteratively and progressively replace the oldest samples in the \textit{same} queue list according to their labels or pseudo-babels. This design tailors ground-truth labels from the labeled data into the shared queue list, thus improving the accuracy of candidate keys for unlabeled queries $\mathbf{q}_i^U$ than that of a separate queue for unlabeled data.

Besides $L_{\text{PGC}} $ and $\widehat{L}_{\text{PGC}}$,  a standard cross-entropy (CE) loss on labeled data is applied on the prediction probability $\mathbf{p}_i = g(f(\mathbf{x}_i^L))$ for each data point $\mathbf{x}_i^L$ as Eq.~\eqref{CE_loss}. The overall loss function of Self-Tuning can be formulated as follows:
$$\mathbb{E}_{ \left( \mathbf { x } _ { i } , {y} _ { i } \right) \in \mathcal{L}} \left(L_{\rm CE} +  L_{\rm PGC}\right) +  \mathbb{E}_{  (\mathbf { x } _ { i })  \in \mathcal{U}} \widehat{L}_{\rm PGC}.$$
It is worthy to mention that \textit{no trade-off coefficients between the above losses are introduced} since the magnitude of these loss terms is comparable. In summary, the network architecture of Self-Tuning is illustrated in Figure~\ref{fig:arch}.

\begin{table*}[htbp]
	\centering
	\captionof{table}{Classification accuracy (\%)  $\uparrow$ of Self-Tuning and various baselines on standard TL benchmarks (ResNet-50 pre-trained).}
	\addtolength{\tabcolsep}{-4pt} 
	\label{table_TL_benchmark}
		\vskip 0.05in
	\begin{tabular}{l|c|l|cccc}
		\toprule
		\multirow{2}*{Dataset} &  \multirow{2}*{Type} & \multirow{2}*{Method} & \multicolumn{4}{c}{Label Proportion} \\
		\cmidrule(lr){4-7}
		& & & $15\%$ & $30\%$ & $50\%$ & $100\%$ \\
		\midrule
		\multirow{15}*{\textit{CUB-200-2011}} & \multirow{5}*{TL} & Fine-Tuning (baseline) &  45.25{\scriptsize $\pm$0.12} & 59.68{\scriptsize $\pm$0.21}  & 70.12{\scriptsize $\pm$0.29} & 78.01{\scriptsize $\pm$0.16} \\
		& & $\rm{L^2}$-$\rm{SP}$ \cite{Li2018L2SP} & 45.08{\scriptsize $\pm$0.19}  & 57.78{\scriptsize $\pm$0.24} & 69.47{\scriptsize $\pm$0.29} & 78.44{\scriptsize $\pm$0.17} \\
		& & $\rm{DELTA}$ \cite{Li2019Delta} & 46.83{\scriptsize $\pm$0.21} & 60.37{\scriptsize $\pm$0.25} & 71.38{\scriptsize $\pm$0.20} & 78.63{\scriptsize $\pm$0.18} \\
		& & $\rm{BSS}$ \cite{Chen2019BSS} & 47.74{\scriptsize $\pm$0.23}  & 63.38{\scriptsize $\pm$0.29}  & 72.56{\scriptsize $\pm$0.17} & 78.85{\scriptsize $\pm$0.31}  \\
		& & Co-Tuning~\cite{you2020co}  & 52.58{\scriptsize $\pm$0.53} & 66.47{\scriptsize $\pm$0.17} & 74.64{\scriptsize $\pm$0.36} &  81.24{\scriptsize $\pm$0.14}  \\
		
		\cmidrule(lr){2-7}
		& \multirow{6}*{SSL} & $\Pi$-model~\cite{Laine17Pimodel} & 45.20{\scriptsize $\pm$0.23} & 56.20{\scriptsize $\pm$0.29} & 64.07{\scriptsize $\pm$0.32} & -- \\
		&& Pseudo-Labeling~\cite{Lee13Pseudolabel} &45.33{\scriptsize $\pm$0.24}& 62.02{\scriptsize $\pm$0.31} & 72.30{\scriptsize $\pm$0.29}  & --\\
		& & Mean Teacher~\cite{Tarvainen17MeanTeacher} &  53.26{\scriptsize $\pm$0.19} & 66.66{\scriptsize $\pm$0.20} & 74.37{\scriptsize $\pm$0.30} & --\\
		& & UDA~\cite{xie2019unsupervised} & 46.90{\scriptsize $\pm$0.31}  & 61.16{\scriptsize $\pm$0.35}  & 71.86{\scriptsize $\pm$0.43} & --\\
		& & FixMatch~\cite{sohn2020fixmatch} & 44.06{\scriptsize $\pm$0.23} & 63.54{\scriptsize $\pm$0.18} & 75.96{\scriptsize $\pm$0.29} & -- \\
		& & SimCLRv2 ~\cite{chen2020big} & 45.74{\scriptsize $\pm$0.15} & 62.70{\scriptsize $\pm$0.24} & 71.01{\scriptsize $\pm$0.34} & -- \\
		\cmidrule(lr){2-7}
		
		& \multirow{3}*{Combine}  & Co-Tuning + Pseudo-Labeling & 54.11{\scriptsize $\pm$0.24} &68.07{\scriptsize $\pm$0.32}& 75.94{\scriptsize $\pm$0.34}  & --\\

		& & Co-Tuning + Mean Teacher & 57.92{\scriptsize $\pm$0.18} & 67.98{\scriptsize $\pm$0.25} & 72.82{\scriptsize $\pm$0.29} & --\\
		
		&& Co-Tuning + FixMatch  &46.81{\scriptsize $\pm$0.21} &58.88{\scriptsize $\pm$0.23}  & 73.07{\scriptsize $\pm$0.29} & --\\
		
		\cmidrule(lr){2-7}
		& & \textbf{Self-Tuning (ours)}& \textbf{64.17}{\scriptsize $\pm$0.47} & \textbf{75.13}{\scriptsize $\pm$0.35} & \textbf{80.22}{\scriptsize $\pm$0.36}& \textbf{83.95}{\scriptsize $\pm$0.18}\\
		
		\midrule
		\multirow{15}*{\textit{Stanford Cars}} & \multirow{5}*{TL} & Fine-Tuning (baseline)& 36.77{\scriptsize $\pm$0.12} & 60.63{\scriptsize $\pm$0.18} & 75.10{\scriptsize $\pm$0.21} & 87.20{\scriptsize $\pm$0.19}\\
		&& $\rm{L^2}$-$\rm{SP}$ \cite{Li2018L2SP} & 36.10{\scriptsize $\pm$0.30}& 60.30{\scriptsize $\pm$0.28}& 75.48{\scriptsize $\pm$0.22}&  86.58{\scriptsize $\pm$0.26}\\
		&& $\rm{DELTA}$ \cite{Li2019Delta} & 39.37{\scriptsize $\pm$0.34}& 63.28{\scriptsize $\pm$0.27} & 76.53{\scriptsize $\pm$0.24}&  86.32{\scriptsize $\pm$0.20}  \\
		&& $\rm{BSS}$ \cite{Chen2019BSS} & 40.57{\scriptsize $\pm$0.12}& 64.13{\scriptsize $\pm$0.18}& 76.78{\scriptsize $\pm$0.21}&  87.63{\scriptsize $\pm$0.27} \\
		&& Co-Tuning~\cite{you2020co} & 46.02{\scriptsize $\pm$0.18}& 69.09{\scriptsize $\pm$0.10}& 80.66{\scriptsize $\pm$0.25}&  89.53{\scriptsize $\pm$0.09} \\
		\cmidrule(lr){2-7}
		& \multirow{6}*{SSL} & $\Pi$-model~\cite{Laine17Pimodel} & 45.19{\scriptsize $\pm$0.21} & 57.29{\scriptsize $\pm$0.26} & 64.18{\scriptsize $\pm$0.29} & --\\
		&& Pseudo-Labeling~\cite{Lee13Pseudolabel}  &40.93{\scriptsize $\pm$0.23} &67.02{\scriptsize $\pm$0.19}& 78.71{\scriptsize $\pm$0.30} & --\\
		&& Mean Teacher~\cite{Tarvainen17MeanTeacher} & 54.28{\scriptsize $\pm$0.14} & 66.02{\scriptsize $\pm$0.21} & 74.24{\scriptsize $\pm$0.23} & -- \\
		&& UDA~\cite{xie2019unsupervised} & 39.90{\scriptsize $\pm$0.43} & 64.16{\scriptsize $\pm$0.40} & 71.86{\scriptsize $\pm$0.56} & --\\
		&& FixMatch~\cite{sohn2020fixmatch} & 49.86{\scriptsize $\pm$0.27}  & 77.54{\scriptsize $\pm$0.29}& 84.78{\scriptsize $\pm$0.33} & --\\
		& &SimCLRv2 ~\cite{chen2020big} & 45.74{\scriptsize $\pm$0.16} & 61.70{\scriptsize $\pm$0.18} & 77.49{\scriptsize $\pm$0.24} & --\\
		\cmidrule(lr){2-7}

			&\multirow{3}*{Combine} & Co-Tuning + Pseudo-Labeling &50.16{\scriptsize $\pm$0.23} &73.76{\scriptsize $\pm$0.26}&83.33{\scriptsize $\pm$0.34} & --\\
	
		& & Co-Tuning + Mean Teacher & 52.98{\scriptsize $\pm$0.19} & 71.42{\scriptsize $\pm$0.24} &  75.38{\scriptsize $\pm$0.29} & --\\
		
			& & Co-Tuning + FixMatch  &42.34{\scriptsize $\pm$0.19} & 73.24{\scriptsize $\pm$0.25}   &  83.13{\scriptsize $\pm$0.34} & --\\
		
		\cmidrule(lr){2-7}
		&& \textbf{Self-Tuning (ours)}  & \textbf{72.50}{\scriptsize $\pm$0.45} & \textbf{83.58}{\scriptsize $\pm$0.28} & \textbf{88.11}{\scriptsize $\pm$0.29}&  \textbf{90.67}{\scriptsize $\pm$0.23} \\
		
		\midrule
		\multirow{15}*{\textit{FGVC Aircraft}} & \multirow{5}*{TL} & Fine-tuning (baseline) & 39.57{\scriptsize $\pm$0.20}& 57.46{\scriptsize $\pm$0.12}& 67.93{\scriptsize $\pm$0.28}&  81.13{\scriptsize $\pm$0.21}\\
		&& $\rm{L^2}$-$\rm{SP}$ \cite{Li2018L2SP} & 39.27{\scriptsize $\pm$0.24}& 57.12{\scriptsize $\pm$0.27}& 67.46{\scriptsize $\pm$0.26}&  80.98{\scriptsize $\pm$0.29}\\
		&& DELTA~\cite{Li2019Delta} & 42.16{\scriptsize $\pm$0.21}& 58.60{\scriptsize $\pm$0.29}& 68.51{\scriptsize $\pm$0.25}&  80.44{\scriptsize $\pm$0.20}\\
		&& BSS \cite{Chen2019BSS} & 40.41{\scriptsize $\pm$0.12}& 59.23{\scriptsize $\pm$0.31}& 69.19{\scriptsize $\pm$0.13}&  81.48{\scriptsize $\pm$0.18}\\
		&& Co-Tuning \cite{you2020co} & 44.09{\scriptsize $\pm$0.67}& 61.65{\scriptsize $\pm$0.32} & 72.73{\scriptsize $\pm$0.08}&  83.87{\scriptsize $\pm$0.09} \\
		\cmidrule(lr){2-7}
		& \multirow{6}*{SSL}  & $\Pi$-model~\cite{Laine17Pimodel} & 37.32{\scriptsize $\pm$0.25} & 58.49{\scriptsize $\pm$0.26} & 65.63{\scriptsize $\pm$0.36} & --\\
		&& Pseudo-Labeling~\cite{Lee13Pseudolabel} &46.83{\scriptsize $\pm$0.30} &62.77{\scriptsize $\pm$0.31}&73.21{\scriptsize $\pm$0.39} & --\\
		&& Mean Teacher~\cite{Tarvainen17MeanTeacher} & 51.59{\scriptsize $\pm$0.23} & 71.62{\scriptsize $\pm$0.29} & 80.31{\scriptsize $\pm$0.32} & --\\
		&& UDA~\cite{xie2019unsupervised} & 43.96{\scriptsize $\pm$0.45}  & 64.17{\scriptsize $\pm$0.49}  & 67.42{\scriptsize $\pm$0.53} & --\\
		&& FixMatch~\cite{sohn2020fixmatch} & 55.53{\scriptsize $\pm$0.26} & 71.35{\scriptsize $\pm$0.35} & 78.34{\scriptsize $\pm$0.43} & --\\
		&& SimCLRv2 ~\cite{chen2020big} & 40.78{\scriptsize $\pm$0.21}  & 59.03{\scriptsize $\pm$0.29}  & 68.54{\scriptsize $\pm$0.30}  & --\\
		\cmidrule(lr){2-7}
		
			&\multirow{3}*{Combine} & Co-Tuning + Pseudo-Labeling &49.15{\scriptsize $\pm$0.32} & 65.62{\scriptsize $\pm$0.34} &74.57{\scriptsize $\pm$0.40} & --\\
		& & Co-Tuning + Mean Teacher & 51.46{\scriptsize $\pm$0.25} & 64.30{\scriptsize $\pm$0.28} & 70.85{\scriptsize $\pm$0.35} & --\\
			& & Co-Tuning + FixMatch  &53.74{\scriptsize $\pm$0.23} &  69.91{\scriptsize $\pm$0.26} &  80.02{\scriptsize $\pm$0.32} & --\\
		
		\cmidrule(lr){2-7}
		
		&& \textbf{Self-Tuning (ours)}  & \textbf{64.11}{\scriptsize $\pm$0.32} & \textbf{76.03}{\scriptsize $\pm$0.25} & \textbf{81.22}{\scriptsize $\pm$0.29} &  \textbf{84.28}{\scriptsize $\pm$0.14}  \\
		\bottomrule
	\end{tabular}
\end{table*}

\section{Experiments}

We empirically evaluate Self-Tuning in several {dimensions}: (1) \textbf{Task Variety}: four visual tasks with various dataset scales including \textit{CUB-200-2011}~\cite{WahCUB_200_2011}, \textit{Stanford Cars}~\cite{KrauseStarkDengFei-Fei_3DRR2013} and \textit{FGVC Aircraft}~\cite{MajiRKBV13Air} and {\textit{CIFAR-100}}~\cite{Krizhevsky09cifar}
	, as well as one NLP task: {CoNLL 2013}~\cite{SangM03CoNLL}.
	(2) \textbf{Label Proportion}: the proportion of labeled dataset \textit{ranging from $15\%$ to $50\%$ following the common practice of transfer learning}, as well as including \textit{$4$ labels and $25$ labels per class following the popular protocol of semi-supervised learning}.
	(3) \textbf{Pre-trained models}: mainstream pre-trained models are adopted including ResNet-18, ResNet-50~\cite{he2016deep}, EfficientNet~\cite{Tan19EfficientNet},  MoCov2~\cite{he2019momentum} and BERT~\citep{devlin2018bert}.
\paragraph{Baselines}
We compared Self-Tuning against three types of {baselines}: (1) \textbf{Transfer Learning (TL)}: besides the vanilla fine-tuning, four
	state-of-the-art TL techniques: {L2SP}~\cite{Li2018L2SP}, {DELTA}~\cite{Li2019Delta}, {BSS}~\cite{Chen2019BSS} and {Co-Tuning}~\cite{you2020co} are included. (2) \textbf{Semi-supervised Learning (SSL)}: we include three classical SSL methods:
	{$\Pi$-model}~\cite{Laine17Pimodel}, {Pseudo-Labeling}~\cite{Lee13Pseudolabel}, and {Mean Teacher}~\cite{Tarvainen17MeanTeacher}, as well as three state-of-the-art SSL methods: {UDA}~\cite{xie2019unsupervised}, {FixMatch}~\cite{sohn2020fixmatch}, and {SimCLRv2}~\cite{chen2020big}.
	 Note that all SSL methods are initialized from a ResNet-50 pre-trained model for a fair comparison with TL methods.
	(3) {\textbf{TL} + \textbf{SSL}}: Strong combinations TL and SSL methods are included as our baselines, including {Co-Tuning} + {FixMatch}, {Co-Tuning} + {Pseudo-Labeling}, {Co-Tuning} + {Mean Teacher}.
	FixMatch, UDA, and Self-Tuning use the same \textit{RandAugment} method, while other baselines use normal ones.


\paragraph{Implementation Details}
For a given pre-trained model, we replace its last-layer with a randomly initialized task-specific layer as the classifier $g$ whose learning rate is $10$ times that for pre-trained parameters, following the common fine-tuning principle~\cite{yosinski2014transferable}. Meanwhile, another randomly initialized projector head $h$ is introduced to generate the representations of the query or key. Following MoCo~\cite{he2019momentum}, we adopted a default temperature $\tau=0.07$, a learning rate ${\rm lr}=0.001$ and a queue size $D=32$ for each category. SGD with a momentum of $0.9$ is adopted as the optimizer. 
Each experiment is repeated three times with different random seeds.
Code will be available at \url{github.com/thuml/Self-Tuning}.

\begin{figure}[htbp]
	\centering
	\subfigure[Acc of FixMatch on \textit{CUB}]{
		\includegraphics[width=0.22\textwidth]{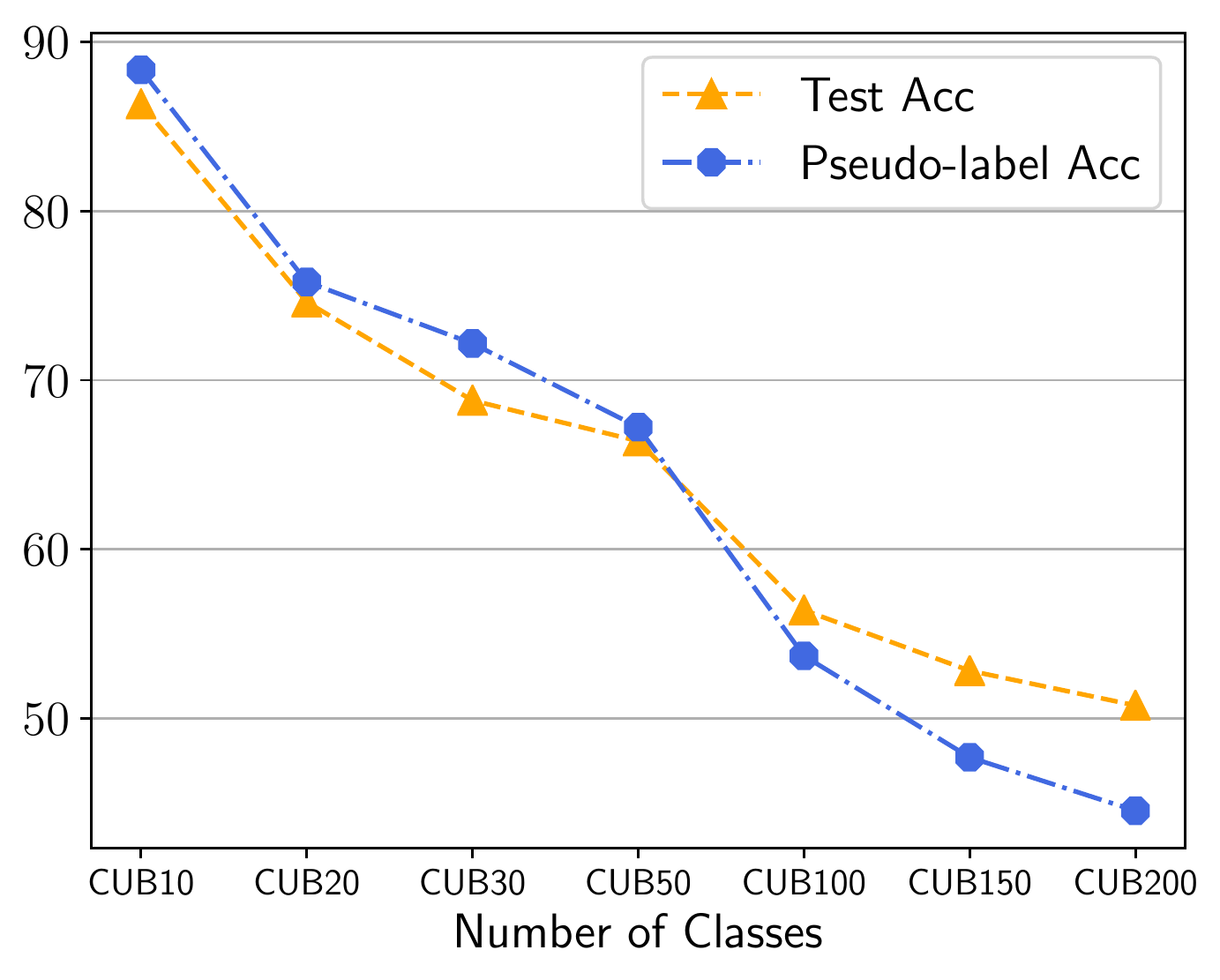}
		\label{fig:label_size}
	}
	\subfigure[Test accuracy of Co-Tuning]{
		\includegraphics[width=0.22\textwidth]{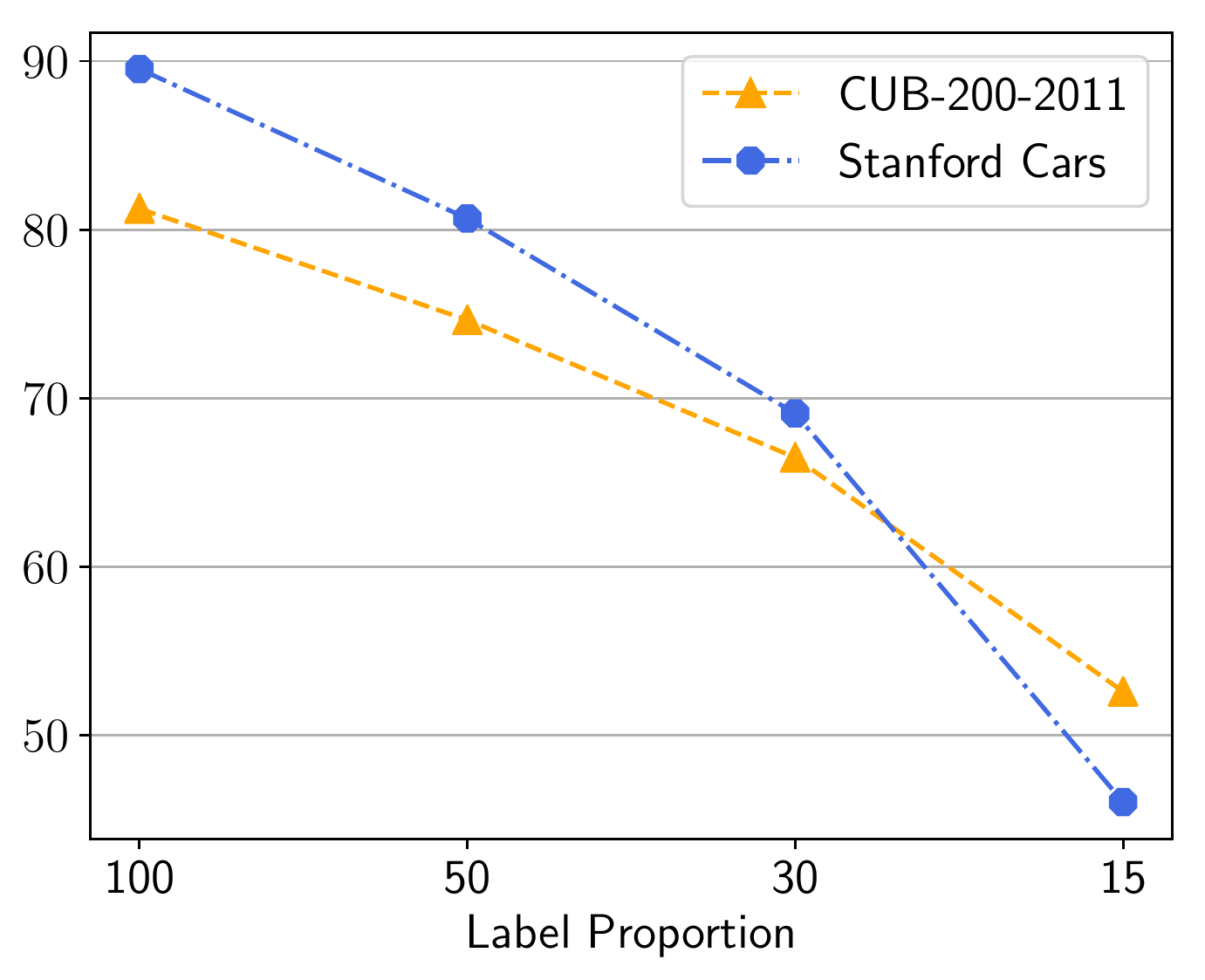}
		\label{fig:label_proportion}
	}
\caption{Test accuracy of a state-of-the-art SSL method and a TL method on various class numbers or label ratios respectively.}
\end{figure}

\subsection{A Prior Study}
In a prior study, we investigated the current state-of-the-art SSL method, FixMatch~\cite{sohn2020fixmatch}, on a target dataset \textit{CUB-200-2011}~\cite{WahCUB_200_2011} containing $200$ bird species. As Figure~\ref{fig:label_size} shows, keeping a same label proportion of 15\%, the test accuracy of FixMatch drops rapidly as the descending accuracy of pseudo-labels when the label space enlarges from $10$ (\textit{CUB10}) to $200$ (\textit{CUB200}). We further investigated the current state-of-the-art TL method,  Co-Tuning, on standard TL benchmarks: \textit{CUB-200-2011}  and \textit{Stanford Cars}~\cite{KrauseStarkDengFei-Fei_3DRR2013}.
As shown in Figure~\ref{fig:label_proportion}, the test accuracy of Co-Tuning declines rapidly as the number of labeled data decreases.  

\subsection{Standard Transfer Learning Benchmarks}
The standard TL benchmarks extensively investigated in previous fine-tuning techniques~\cite{you2020co} consist of  \textit{CUB-200-2011} ($11,788$ images for $200$ bird species), \textit{Stanford Cars} ($16,185$ images for $196$ car categories), and \textit{FGVC Aircraft} ($10,000$ images for $100$ aircraft variants). Co-Tuning has two steps, in which the first step of calculating the category relationship relies on the \textit{certainty} of data augmentation. From Pseudo-Labeling to FixMatch, the randomness of data augmentation increases. Therefore, Pseudo-Labeling benefits from adding Co-Tuning while FixMatch does not, as reported in Table~\ref{table_TL_benchmark}. Further, these results show that neither \textit{a simple combination} of SSL and TL methods nor \textit{a sequential form} between labeled and unlabeled data proposed by a prior work called SimCLRv2 achieve satisfactory performance on the target dataset. 

Contrarily, by unifying the exploration of labeled and unlabeled data and the transfer of a pre-trained model, Self-Tuning outperforms its SSL and TL counterparts by sharp margins across various datasets and different label proportions, \emph{e.g.}~it doubles the accuracy of fine-tuning on \emph{Cars} with $15\%$ labels.
Meanwhile, with only a half of labeled data, Self-Tuning surpasses the fine-tuning method with full labels.
It is noteworthy that Self-Tuning is pretty robust to hyper-parameters: cross-validated on one task works well for these three datasets and label proportions. Further, if the target dataset is fully labeled, Self-Tuning seamlessly boils down to a competitive transfer learning method, as shown in the last column of Table~\ref{table_TL_benchmark}.

%
%
%

\begin{figure}[htbp]
	\centering
	\subfigure[Compare with SupCL]{
		\includegraphics[width=0.22\textwidth]{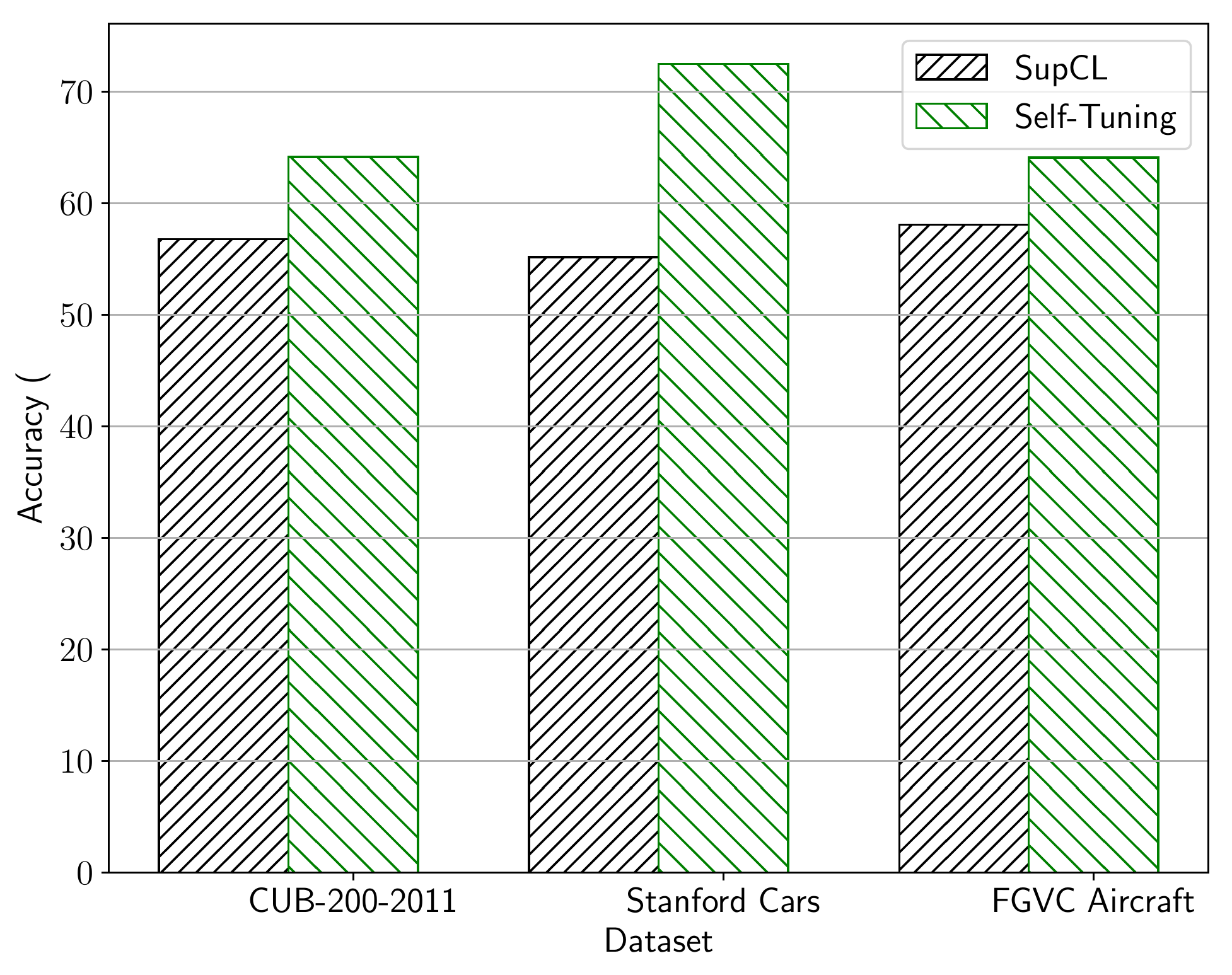}
		\label{fig:SupCL}
	}
	\subfigure[Compare with SimCLRv2]{
		\includegraphics[width=0.22\textwidth]{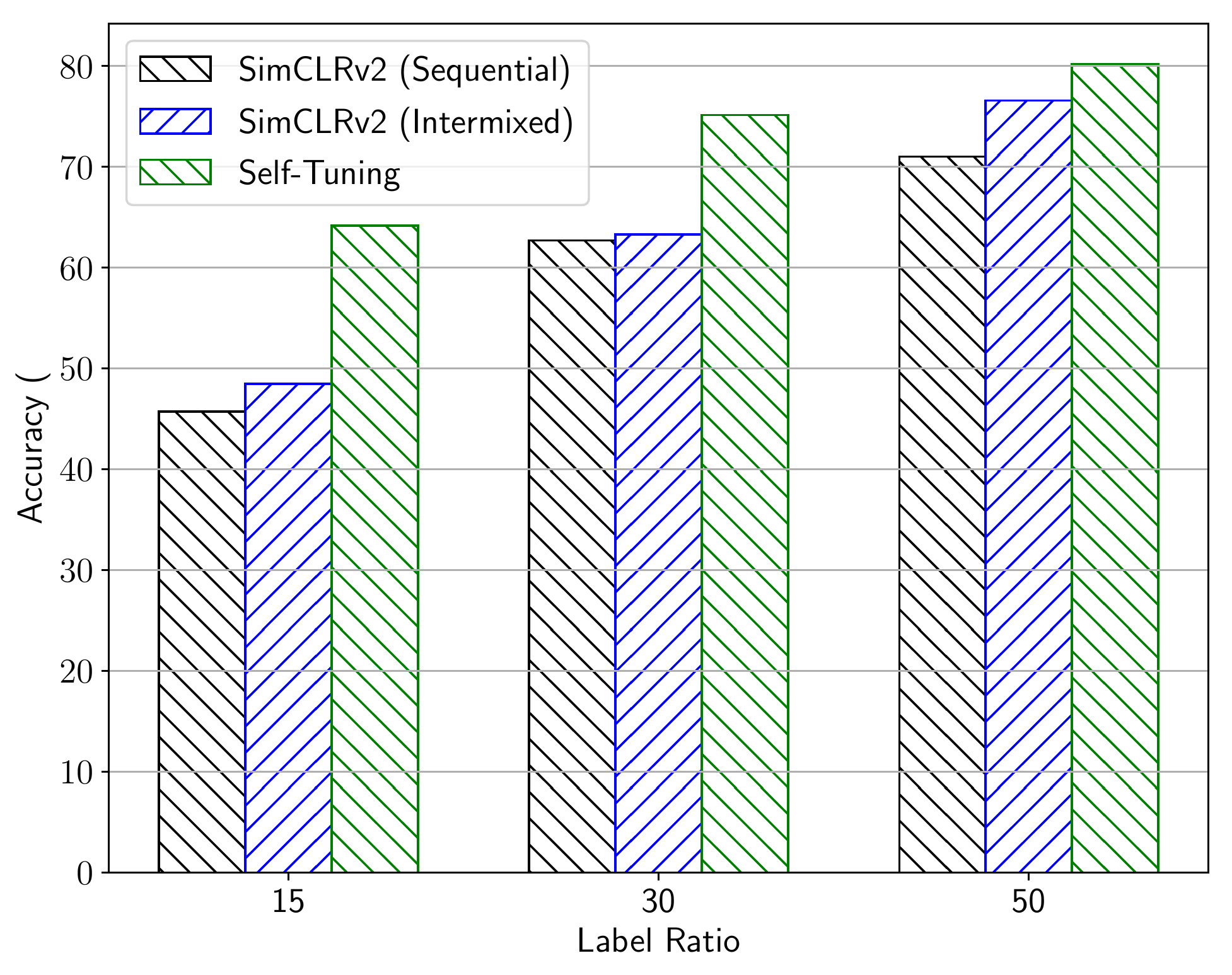}
		\label{fig:simclrv2}
	}
	\caption{The classification accuracy of various methods when comparing the proposed Self-Tuning with SupCL and SimCLRv2: (a) Compare with SupCL on various datasets provided with a label ratio of $15\%$; (b) Compare with SimCLRv2 on \textit{CUB-200-2011} of various label ratios: $15\%$, $30\%$ and $50\%$.}
\end{figure}

\paragraph{Discussion: Compare with SupCL}
A recent method named SupCL (Supervised Contrastive Learning)~\cite{SupCL20NIPS} has a similar equation form with the proposed PGC loss. However, they are different from the following perspectives: 
{(1)} Self-Tuning aims at tackling confirmation bias and model shift issues simultaneously in an efficient one-stage framework while SupCL is designed for pre-training. {(2)} The shared key sets between labeled and unlabeled data enable a unified exploration while SupCL is only for labeled data. {(3)} The {positive and negative size} for each class of Self-Tuning are \textit{fixed and balanced} while those of SupCL are random, making Self-Tuning more robust to imbalanced datasets as shown in Figure~\ref{fig:SupCL}.

\paragraph{Discussion: Compare with SimCLRv2}
In Section~\ref{intro}, we hypothesize that the \textit{sequential} form between first fine-tuning on $\mathcal{L}$ and then distilling on $\mathcal{U}$ that SimCLRv2 adopts
is to blame since the fine-tuned model would easily shift towards the limited labeled data with sampling bias and leaves away from the original smooth model pre-trained on a large-scale dataset. Here, an intuitive idea is to change the sequential form of SimCLRv2 into an intermixed version. As shown in Figure~\ref{fig:simclrv2}, we compare Self-Tuning to SimCLRv2 and see an obvious improvement of the intermixed form over the sequential form. However, both forms of SimCLRv2 still much worse than Self-Tuning.
%

\begin{table}[h]
	\centering
		\addtolength{\tabcolsep}{1pt} 
	\caption{Error rates (\%) $\downarrow$ on standard SSL benchmark: \textit{\textit{CIFAR-100}} provided with only $400$ labels, $2500$ labels and $10000$ labels.}
	\label{cifar100}
	\vskip 0.05in
	\begin{tabular}{l|c|cc}
		\toprule
		Method &  Network &$2.5$k & $10$k \\
		\midrule
		{$\Pi$-Model} & \multirow{5}*{WRN-28-8} & 57.25 & 37.88 \\
		Pseudo-Labeling&&57.38 & 36.21 \\
		Mean Teacher &   \multirow{5}*{\#Para: 11.76M}& 53.91& 35.83 \\
		MixMatch & & 39.94 & 28.31\\
		UDA && 33.13 & 24.50 \\
		ReMixMatch && {27.43}& 23.03 \\
		FixMatch && {28.64} & 23.18 \\
		\midrule
		FixMatch &  \multirow{3}*{EfficientNet-B2}    & 29.99  & 21.69   \\
		{Fine-Tuning} & & 31.69  & 21.74  \\
		Co-Tuning & \multirow{2}*{\#Para: 9.43M}&  30.94  & 22.22   \\
		
		{\textbf{Self-Tuning}}   &&  \textbf{24.16} & {\textbf{17.57}} \\

		\bottomrule
	\end{tabular}
\end{table}

\begin{table}[!h]
	\footnotesize
	\addtolength{\tabcolsep}{-4pt} 
	\caption{Error rates (\%) $\downarrow$ on \textit{\textit{CIFAR-100}} provided with only $400$ labels and a pre-trained EfficientNet-B2 model (CT: Co-Tuning; PL: PseudoLabel; MT: MeanTeacher; FM: FixMatch.)}
	\label{cifar100_appendix}
	\vskip 0.05in
	\begin{tabular}{c|c|c|c|c}
		\toprule
		Fine-Tuning  & L2SP  & DELTA & BSS & Co-Tuning \\
		\midrule
		60.79 & 59.21 & 58.23 & 58.49  & 57.58\\
		\midrule
		$\Pi$-model  & PseudoLabel  & MeanTeacher & FixMatch  & UDA \\
		\midrule
		60.50 & 59.21 & 60.68 & 57.87  & 58.32 \\
		\midrule
		SimCLRv2 & CT+PL  & CT+MT  & CT+FM & Self-Tuning  \\
		\midrule
		59.45 & 56.21 & 56.78 & 57.94 &{ 47.17}  \\
		\bottomrule
	\end{tabular}
\end{table}

\subsection{Standard Semi-supervised Learning Benchmarks}
 We adopt the most difficult \textit{CIFAR-100} dataset with $100$ categories among the famous SSL benchmarks including \textit{CIFAR-100}, \textit{CIFAR-10}, \textit{SVHN}, and \textit{STL-10}, where the last three datasets have only $10$ categories. 
Since a WRN-28-8~\cite{ZagoruykoK16WideResNet} model pre-trained on ImageNet is not openly available, we adopt an EfficientNet-B2 model with much fewer parameters instead. As shown in Table~\ref{cifar100} and Table~\ref{cifar100_appendix},  FixMatch works worse on EfficientNet-B2 than on WRN-28-8, while Self-Tuning outperforms the strongest baselines on WRN-28-8 by large margins. For a fair comparison, we further provided all baselines on EfficientNet-B2 to verify the superiority of Self-Tuning.


\begin{table}[htbp]
	\addtolength{\tabcolsep}{-4.5pt} 
	\caption{Classification accuracy (\%) $\uparrow$ with a typical unsupervised pre-trained model MoCov2 on \textit{CUB-200-2011}.}
	\label{extend_results}
	\centering
		\vskip 0.05in
	\begin{tabular}{l|l|cc}
		\toprule
		
		Type & Method  & $800$ labels  & $5$k labels  \\
		\midrule
		\multirow{2}*{TL}  & Fine-Tuning (baseline) & 20.04 &  71.50  \\
		& Co-Tuning & 20.99 & 71.61  \\
		
		\midrule
		\multirow{2}*{SSL}  & Mean Teacher& 28.13 & 71.26  \\
		& FixMatch &21.18  & 71.28  \\
		
		\midrule
		\multirow{2}*{Combine}   & Co-Tuning + Mean Teacher & 28.43  & 72.21 \\
		& Co-Tuning + FixMatch  & 21.08 	& 71.40   \\
		
		\midrule
		& \textbf{Self-Tuning (ours)}& \textbf{36.80} & \textbf{74.56}  \\
		\bottomrule
	\end{tabular}
\end{table}

\begin{table}[htbp]
	\caption{Ablation studies of Self-Tuning on \emph{Stanford Cars}.}
	\label{ablation_study}
		\vskip 0.05in
	\begin{tabular}{l|l|cc}
		\toprule
		Perspective &  Method & 15\% & 30\%  \\
		\midrule
		\multirow{3}*{Loss Function} & w/ CE loss & 40.93 & 67.02 \\
		&  w/ CL loss & 46.29 & 68.82 \\
		&  w/ PGC loss & \textbf{72.50} & \textbf{83.58} \\
		
		\cmidrule(lr){1-4}
		\multirow{5}*{Info. Exploration} & w/o $\widehat{L}_{\text{PGC}} $& 58.82 & 81.71 \\
		& w/o $L_{\text{PGC}} $ & 58.85 & 77.52  \\ 
		& separate queue & 70.43 & 80.78 \\
		& unified exploration  & \textbf{72.50} & \textbf{83.58}\\
		\bottomrule
	\end{tabular}
\end{table}

\subsection{Unsupervised Pre-trained Models}
Besides initializing from supervised pre-trained models, we further explore the performance of Self-Tuning transferring on an unsupervised pre-trained model named MoCov2~\cite{he2019momentum}. 
As reported in Table~\ref{extend_results},  Self-Tuning yields consistent gains over SSL and TL methods, revealing that Self-Tuning is not bound to specific pre-trained pretext tasks.

\subsection{Named Entity Recognition}
We conduct experiments on CoNLL 2003~\cite{SangM03CoNLL}, an English named entity recognition (NER) task as a token-level classification problem, to explore the performance of Self-Tuning on NLP tasks.  Following the protocol of Co-Tuning, we also adopt BERT~\cite{devlin2018bert} as the pre-trained model (masked language modeling one). Measured by the F1-score of named entities, the vanilla fine-tuning baseline achieves an F1-score of $90.81$, BSS, L2-SP and Co-Tuning achieve $90.85$, $91.02$ and $91.27$ respectively, while Self-Tuning achieves a new state-of-the-art of $94.53$. 

\subsection{Ablation Studies}
We conduct ablation studies in Table~\ref{ablation_study} from two perspectives: (a) Loss Function Type: the assumption in Section~\ref{PGC_section} that PGC loss is much better than CE loss and CL loss for data-efficient deep learning is empirically verified here. (b) Information Exploration Type: by comparing Self-Tuning with models without PGC loss on $\mathcal{L}$ or $\mathcal{U}$, and a model with separate queue lists for $\mathcal{L}$ and $\mathcal{U}$, we demonstrate that the unified exploration is the best choice.

\subsection{Sensitivity Analysis}
\begin{figure}[!h]
	\centering
	\subfigure[Acc on \textit{Car} with $15\%$ labels]{
		\includegraphics[width=0.22\textwidth]{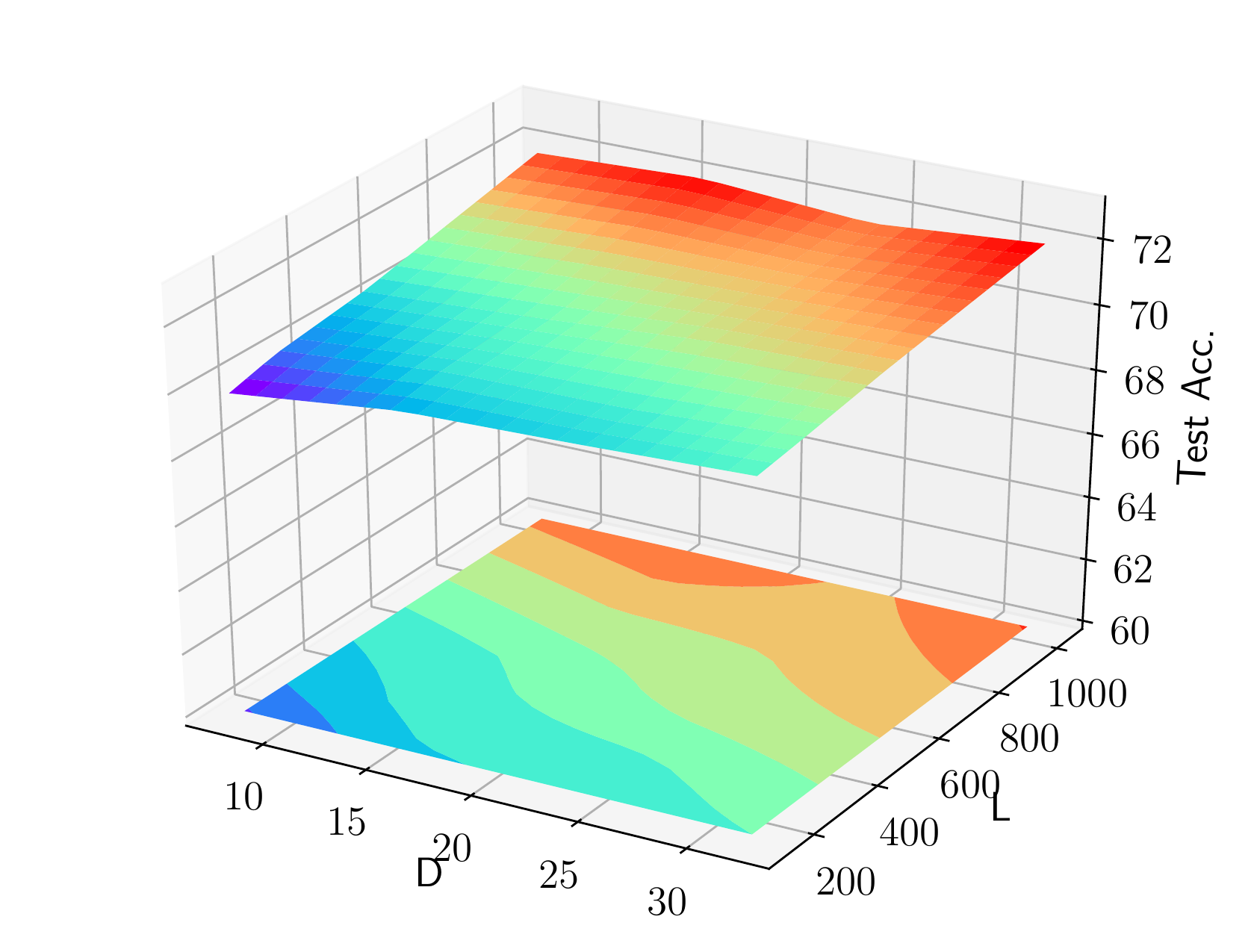}
		\label{fig:sensitivity_car15}
	}
	\subfigure[Acc on \textit{Car} with $30\%$ labels]{
		\includegraphics[width=0.22\textwidth]{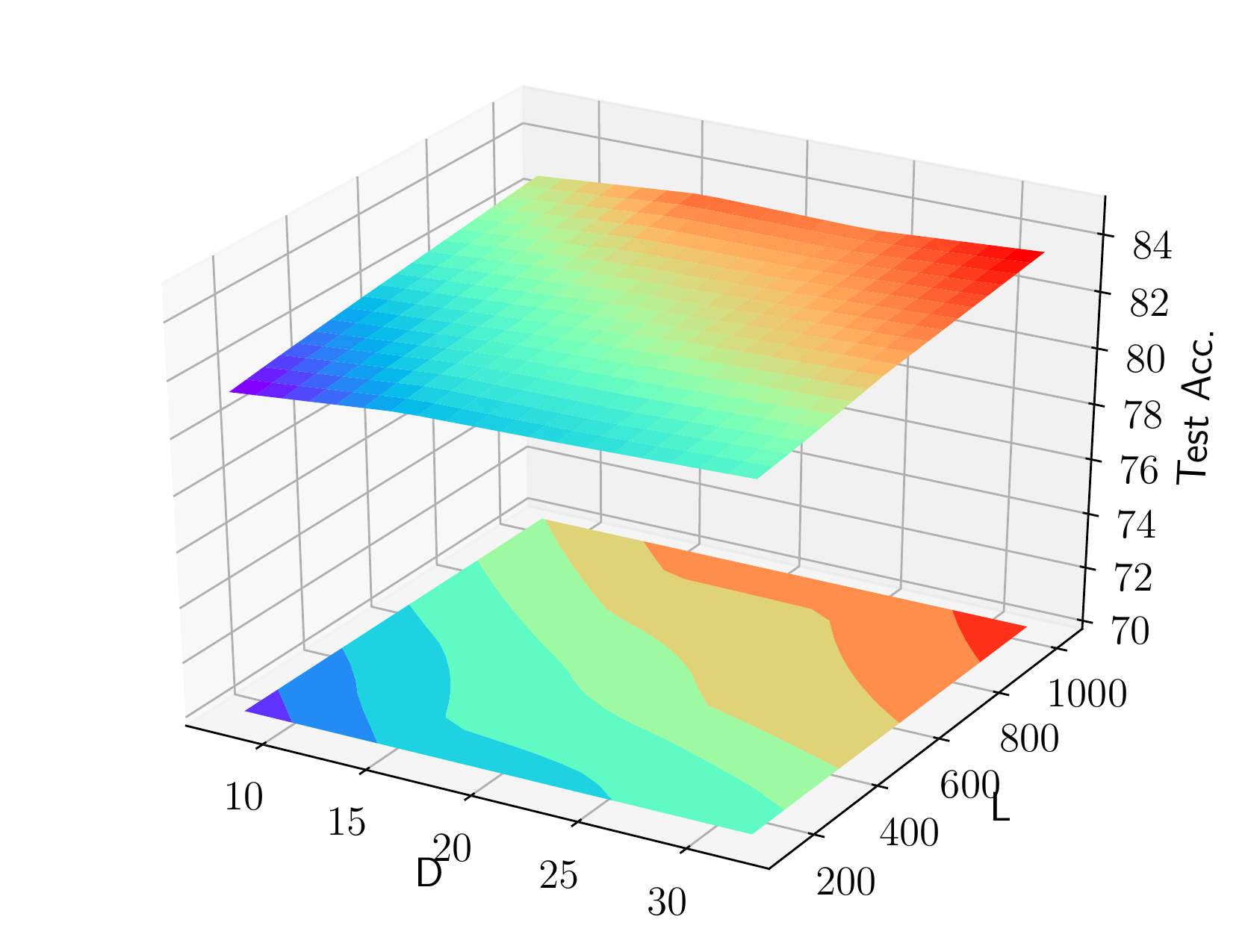}
		\label{fig:sensitivity_car30}
	}
	\caption{Sensitivity analysis for embedded size $L$ of the projector and queue size $D$ of each class on \textit{Stanford Cars}. (Warmer colors indicate higher values)}
	\label{sensitivity_analysis}
\end{figure}

Different from most self-training methods, Self-Tuning is \textit{free of confidence-threshold hyper-parameter $t$ and trade-off coefficients between various losses}. However, it still has two hyper-parameters: feature size $L$ of the projector head $h$ and queue size $D$ for each category, by introducing a pseudo group contrast mechanism. As shown in Figure~\ref{sensitivity_analysis}, Self-Tuning is robust to different values of $L$ and $D$ but tends to prefer larger values of them.

\begin{figure}[!h]
	\centering
	\subfigure[Training Process on \textit{CUB30}]{
		\includegraphics[width=0.22\textwidth]{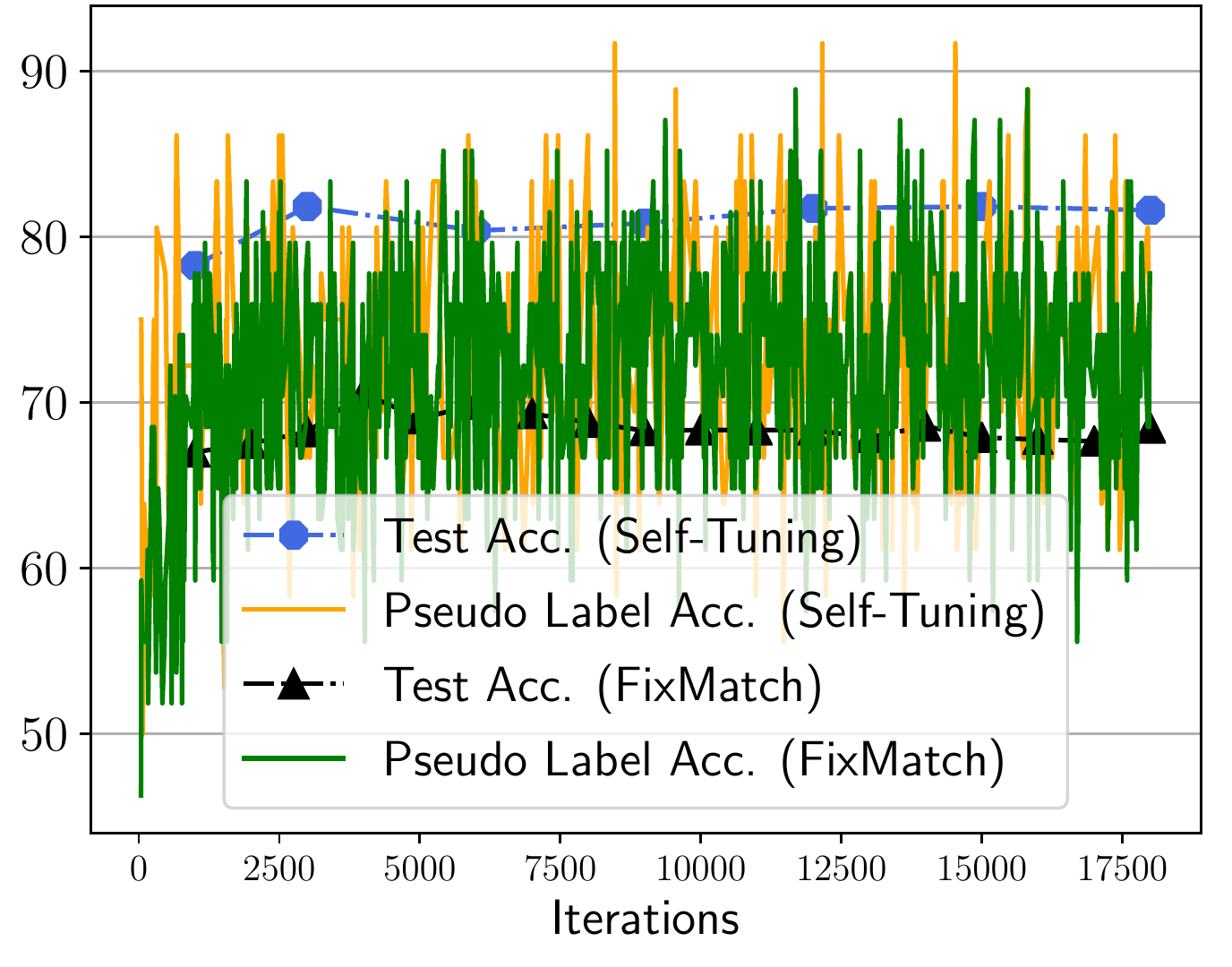}
		\label{fig:training_process}
	}
	\subfigure[${\rm Acc}_{\rm test} - {\rm Acc}_{\rm pseudo\_labels}$]{
		\includegraphics[width=0.22\textwidth]{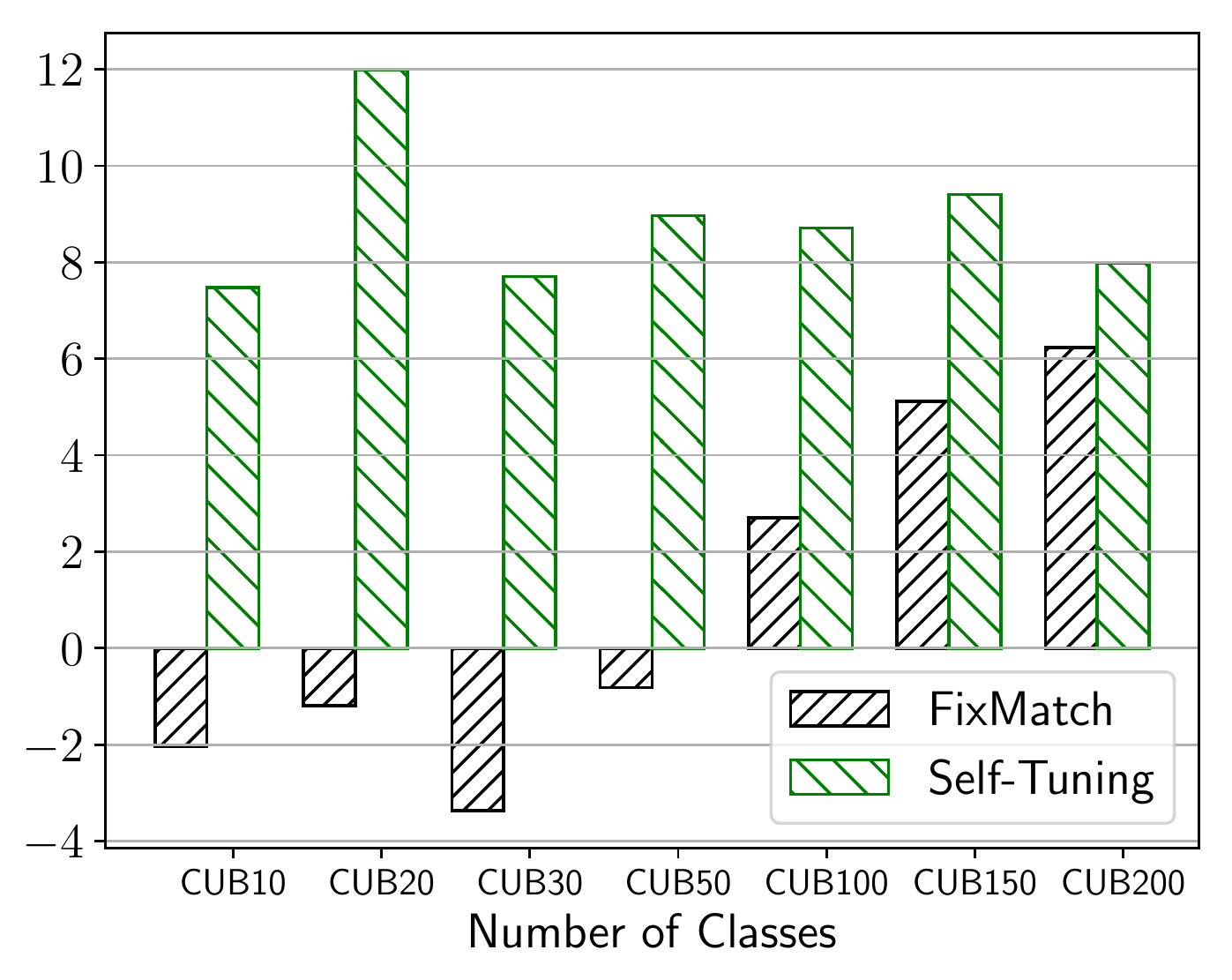}
		\label{fig:accuracy_improvement}
	}
	\caption{Comparisons between Self-Tuning with FixMatch on pseudo label accuracy and test accuracy.}
	\label{comfirmation_bias}
\end{figure}

\subsection{Why Self-Tuning Works}
First, by unifying the exploration of labeled and unlabeled data and the transfer of a pre-trained model, Self-Tuning escapes from the dilemma of just developing TL or SSL methods. Further, Figure~\ref{comfirmation_bias} reveals that the proposed PGC mechanism successfully boosts the tolerance to false labels, since Self-Tuning has a larger improvement over the accuracy of pseudo-labels than FixMatch, given an identical pre-trained model with approximate pseudo-label accuracy.

\section{Conclusion}

Mitigating the requirement for labeled data is a vital issue in deep learning community. However, common practices of TL and SSL only focus on either the pre-trained model or unlabeled data. This paper unleashes the power of both of them by proposing \textit{a new setup} named data-efficient deep learning. To address the challenge of confirmation bias in self-training, a general \textit{Pseudo Group Contrast} mechanism is devised to mitigate the reliance on pseudo-labels and boost the tolerance to false labels. To tackle the model shift problem, we unify the exploration of labeled and unlabeled data and the transfer of a pre-trained model, with a shared key queue beyond just `parallel training'. 

%

\section*{Acknowledgements}
This work was kindly supported by the National Key R\&D Program of China (2020AAA0109201), NSFC grants (62022050, 62021002, 61772299), Beijing Nova Program (Z201100006820041), and MOE Innovation Plan of China.

%
%
%


\bibliography{citations}
\bibliographystyle{icml2021}

%
%
%

\end{document}